%% file: main.tex
\documentclass[conference]{IEEEtran}
\IEEEoverridecommandlockouts

\usepackage{cite}
\usepackage{amsmath,amssymb,amsfonts}
\usepackage{graphicx}
\usepackage{textcomp}
\usepackage[dvipsnames]{xcolor}
\usepackage{url}
\usepackage[hidelinks,colorlinks=false,pdfborder={0 0 0}]{hyperref}

\graphicspath{{figures/}}
\input{macros}

\title{\LARGE \bf IDDMBSE: Integrating Data-Driven and Model-Based Systems Engineering for Trusted Autonomous Cyber-Physical Systems}

\author{John S. Baras, Sai Sandeep Damera, Ryan Matheu, Clinton Enwerem, Praveen M.S. Kumar%
\thanks{This work was partially supported by the Army Research Laboratory Cooperative Agreement No. W911NF-23-2-0040, by a Northrop Grumman Corporation grant, the Lockheed Martin Chair in Systems Engineering, and by the NVIDIA Academic Grant Program using NVIDIA RTX PRO 6000 Blackwell GPUs.}%
\thanks{The authors are with the Institute for Systems Research, University of Maryland, College Park, MD, USA. {\tt\small \{baras, sdamera, rmatheu, enwerem, praveenm\}@umd.edu}.}%
}

\begin{document}
\maketitle
\thispagestyle{empty}
\pagestyle{empty}

\begin{abstract}
Autonomous cyber-physical systems (CPS) sit at the intersection of Model-Based Systems Engineering (MBSE) and data-driven Machine Learning and Artificial Intelligence (ML/AI), yet no integrated Systems Engineering (SE) methodology natively spans both. We address this gap with \iddmbse{}, an Integrated Data-Driven and Model-Based Systems Engineering methodology that extends the rigorous MBSE V-process with a data-driven loop at every step, anchored in SysML, the autonomy stack, and a hybrid model-based plus data-driven trade-off architecture. We instantiate \iddmbse{} as an interoperable, open-source tool chain: \perfect{}, which maps SysML system architectures to executable ROS autonomy stacks for scalable performance evaluation; \tradesx{}, which decomposes design-space exploration into a model-based optimization stage followed by a data-driven evaluation stage; and \veritas{}, which combines formal, data-driven, and runtime verification into a single assurance workflow. We demonstrate \iddmbse{} on a Trusted Autonomous Ground Robot across its development lifecycle, spanning sensor-suite selection, risk-sensitive path planning, behavior-tree task verification, conformal-prediction-based robust perception, and assured multi-robot coordination, all exercised in a contested-terrain Isaac Sim test range that we release with the tool chain. We close by sketching how \iddmbse{} is being re-formulated on SysML\,v2 / KerML foundations to enable language-native composability and tighter ML/AI integration.
\end{abstract}

\begin{IEEEkeywords}
Model-Based Systems Engineering, SysML, SysML\,v2, Trusted Autonomy, Cyber-Physical Systems, Trade-off Analysis, Verification \& Validation.
\end{IEEEkeywords}

\input{sections/introduction}

\input{sections/related_work}
\input{sections/framework}
\input{sections/case_studies}
\input{sections/conclusion}

\section*{Acknowledgment}
The authors thank the U.S.\ Army Research Laboratory for their support and engagement, and Dr. Daniel Hunter and Kashif Ansari for their contributions to this project.

\bibliographystyle{mybibstyle}
\bibliography{references}

\end{document}

%% file: macros.tex
\usepackage{float}
\usepackage{subfig}
\usepackage{booktabs}

\newcommand{\cvar}{\ensuremath{\operatorname{CVaR}}}

\newcommand{\perfect}{\textsc{Perfect}}
\newcommand{\tradesx}{\textsc{Trades-X}}
\newcommand{\veritas}{\textsc{Veritas}}
\newcommand{\iddmbse}{\textsc{IDDMBSE}}

\definecolor{perfectPurple}{HTML}{7C3AED}
\definecolor{tradesBlue}{HTML}{1D4ED8}
\definecolor{veritasGreen}{HTML}{059669}

%% file: sections/introduction.tex
\section{Introduction}
\label{sec:intro}

A modern autonomous robot interleaves model-based components (kinematics, planners, controllers) with learned ones (perception networks, learned policies). This hybrid architecture is the defining feature of the autonomous Cyber-Physical Systems (CPS) now pervasive across transportation, manufacturing, defense, and healthcare, and engineering it holistically and at scale is the open challenge. Yet the the Systems Engineering (SE) community lacks an end-to-end methodology that natively accommodates the data-driven Machine Learning (ML) and Artificial Intelligence (AI) components that dominate modern autonomy stacks. Industrial practice still treats ML/AI subsystems as black boxes plugged into a model-based scaffold, with architecture and assurance arguments retrofitted after the fact. This produces brittle systems, slow design cycles, and weak certification stories, exactly when the demand for \emph{Trusted Autonomy} (autonomy that self-monitors, self-adjusts, and self-learns under verifiable safety guarantees) has become acute.

\begin{figure}[h]
    \centering
    \includegraphics[width=\linewidth]{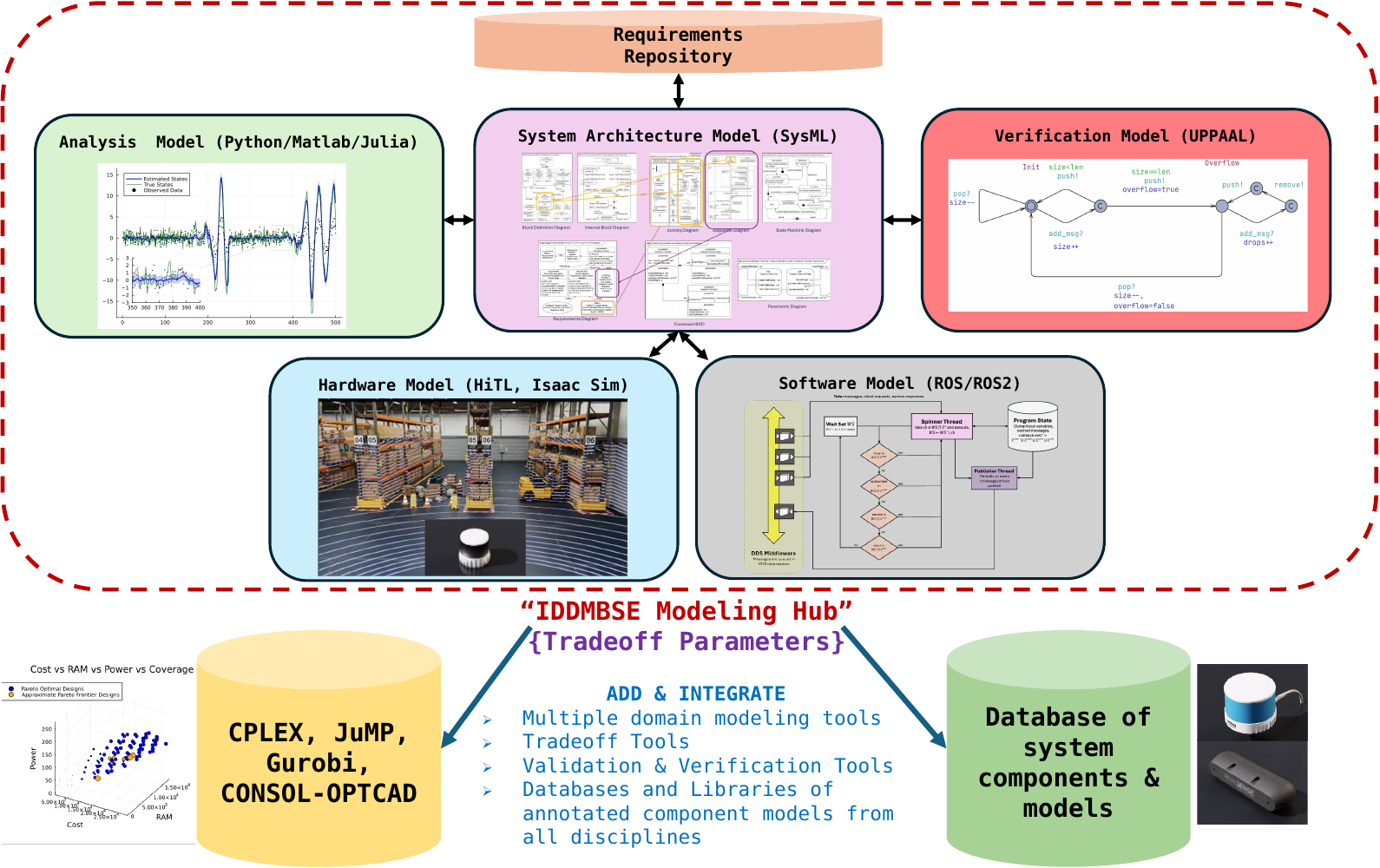}
    \caption{The \iddmbse{} modeling hub. A requirements repository drives a central SysML system-architecture model, augmented by hybrid model instances and coupled to trade-off solvers and a component-model database. In this work, we propose novel tools that realize the hardware/software evaluation, trade-off exploration, and verification regions respectively.}
    \label{fig:iddmbse_framework_hub}
\end{figure}
The root cause is methodological, not technological. Model-Based Systems Engineering (MBSE) and its SysML-anchored toolchain have matured over two decades into a rigorous discipline for designing complex CPS~\cite{friedenthal2014practical,incoseHandbook2023}. It does not, however, scale to autonomy on its own: system and mission complexity grows super-exponentially, and the trustworthy simulations and field experiments needed to certify learned components are rarely available at the scale a purely model-based approach would demand. In parallel, ML/AI has matured into a powerful tool for inferring behaviors from data when first-principles models are unavailable or intractable. These two paradigms answer complementary questions, MBSE asks ``what is the system?'' while ML/AI asks ``what does the system look like in the wild?'', and a usable methodology for autonomous CPS must integrate both without sacrificing the formal traceability MBSE provides or the empirical coverage ML/AI provides. This is the gap addressed by what we have called \textbf{Integrated Data-Driven and Model-Based Systems Engineering (\iddmbse{})}, a natural evolution of the rigorous MBSE V-process in which an augmented data-driven loop is layered on every step (structure, behavior, mapping, requirements, trade-off, verification-validation). The loop is requirements-driven: as the hub in Fig.~\ref{fig:iddmbse_framework_hub} shows, every step begins from the requirements repository, and data-driven effort is spent where the model-based prior is least informative about requirement satisfaction.

\begin{figure}[t]
    \centering
    \includegraphics[width=\linewidth]{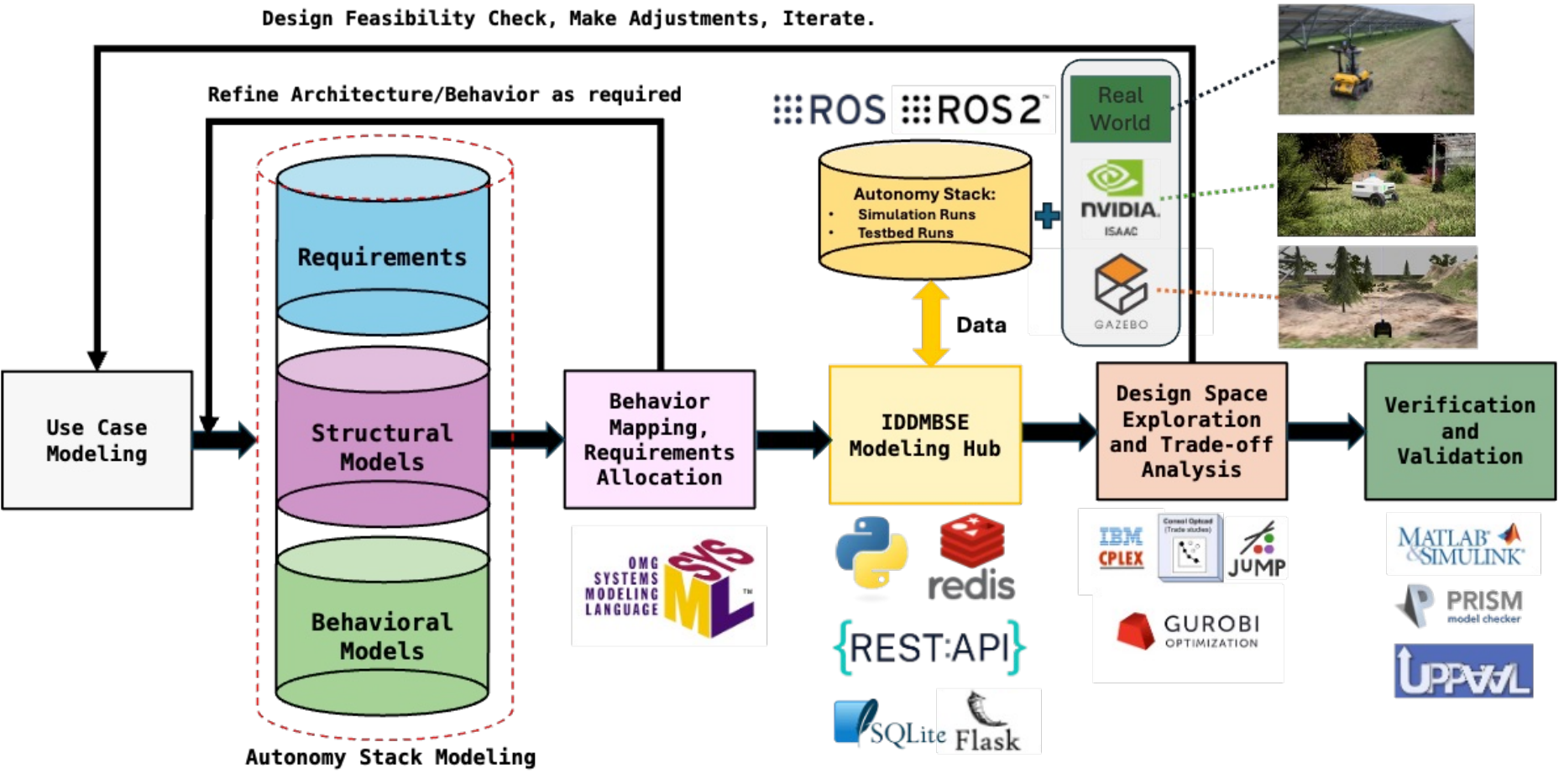}
    \caption{\iddmbse{} extends the rigorous MBSE V-process. SysML serves as the central representation; \perfect{}, \tradesx{}, and \veritas{} provide data-driven augmentation for performance evaluation, design-space exploration, and verification respectively.}
    \label{fig:framework_iddmbse}
\end{figure}

\noindent\textbf{Contributions.} This paper makes three claims, each evaluated in the body and the case study.
\begin{enumerate}
    \item \textbf{Methodology.} We extend the MBSE V-process with a data-driven loop, driven by requirements, and the autonomy stack as a first-class architectural element, yielding a methodology for autonomous CPS that traces ML/AI components, system structure, and requirements through a unified SysML representation.
    \item \textbf{Tool chain.} We instantiate the methodology as three interoperable tools\footnote{The \iddmbse{} tool chain and contested-terrain test range will be released at \url{https://github.com/seil-umd/IDDMBSE}.}, \perfect{} for SysML-to-ROS mapping and scalable performance evaluation, \tradesx{} for hybrid model-based / data-driven design-space exploration, and \veritas{} for combined formal, data-driven, and runtime verification, all coupled through a shared SysML modeling hub.
    \item \textbf{Demonstration.} We exercise the tool chain across the robot development lifecycle on a Trusted Autonomous Ground Vehicle: hardware design (sensor-suite selection with \tradesx{}), software development (path-planning module evaluation with \perfect{}), and behavior assurance (timing and safety verification with \veritas{}), each traced to the same SysML requirements.
\end{enumerate}

The remainder of the paper is organized as follows. Section~\ref{sec:related} situates \iddmbse{} relative to MBSE and Digital Engineering practice, recent AI-for-SE work, and modern V\&V techniques for autonomous CPS. Section~\ref{sec:framework} presents the \iddmbse{} methodology and the three-tool implementation. Section~\ref{sec:case_studies} demonstrates the framework across three robotics use cases: hardware design, path planning, and behavior verification. Section~\ref{sec:conclusion} concludes and outlines how SysML\,v2 and agentic AI will extend the framework.

%% file: sections/related_work.tex
\section{Related Work}
\label{sec:related}

We position \iddmbse{} against three strands of work, using an autonomous ground robot as the running example throughout: model-based systems engineering and the digital-engineering paradigm it anchors; the fast-growing ``AI for systems engineering'' literature that injects ML and AI into the SE workflow; and trade-off analysis and verification for autonomous CPS. Each strand addresses part of the problem, yet none, on its own, makes ML/AI components first-class, requirements-driven citizens of the V-process. Table~\ref{tab:related} makes the comparison precise.

\begin{table}[t]
\centering
\caption{Positioning of \iddmbse{} against related work.}
\label{tab:related}
\footnotesize
\setlength{\tabcolsep}{3pt}
\begin{tabular}{@{}lccccc@{}}
\toprule
 & SysML & Data & ML/AI & Formal & Req.\\
Approach & hub & driven & native & Methods & driven\\
\midrule
Classical MBSE / DE~\cite{friedenthal2014practical,incoseHandbook2023} & \checkmark & & & & \checkmark\\
Hardware sensor design~\cite{frey2025boxi,Milojevic2025CODEI:} & & \checkmark & & & \\
MBSE-driven DSE~\cite{specking2018dse,Harbin2022Model-driven} & \checkmark & \checkmark & & & \\
DE + digital-twin V\&V~\cite{samak-mbse-rdt2025,coleman2025model} & \checkmark & \checkmark & & & \checkmark\\
Robotics benchmarking~\cite{mayoral2024robotperf} & & \checkmark & & & \\
\textbf{\iddmbse{} (ours)} & \checkmark & \checkmark & \checkmark & \checkmark & \checkmark\\
\bottomrule
\end{tabular}
\end{table}

\subsection{MBSE and the Rigorous Digital Engineering Paradigm}
MBSE replaces document-centric SE with executable, traceable models of system structure, behavior, requirements, and parametrics~\cite{friedenthal2014practical}. Its modern formulation is codified in the INCOSE Systems Engineering Handbook~\cite{incoseHandbook2023}, the FuSE (Future of Systems Engineering) initiative~\cite{incoseFuSE}, and the 2018 U.S.\ DoD Digital Engineering (DE) Strategy~\cite{dodDEStrategy}, which together promote a digital-thread-and-twin paradigm across the system life cycle. SysML\,v1 remains the dominant modeling language, while SysML\,v2~\cite{omgSysMLv2} is now stabilizing as a textual, KerML-grounded reformulation with strong composability semantics. Our prior work on MBSE tool integration for robotic manipulation~\cite{meehan2021slip,sekar2022model} and on architecture frameworks for networked CPS~\cite{baras2014fresh,zhou2013cps,spyropoulos2013extending,wang2012integrated} established the SysML hub, trade-off engine, and component-library triad that \iddmbse{} now extends with data-driven components as shown in Figure~\ref{fig:framework_iddmbse}. This lineage gives \iddmbse{} its SysML-based architecture modeling hub and its requirements-first discipline; what the paradigm still lacks is a principled way to fold learned components into that discipline, which is exactly the gap we target here.

\subsection{AI for SE and SE for AI}
The growing literature on AI-for-SE includes generative-AI assistants for SysML modeling~\cite{patelMBSEGenAI2024} and broader MBSE-AI frameworks targeting fast model-driven decisions. On the SE-for-AI side, knowledge-graph and large-language-model based reasoning is increasingly proposed as a complement to model checking in SE workflows~\cite{sun2024thinkongraph}. These works assist a single authoring or reasoning step and stop at the model boundary. None of them orchestrates real-time, distributed simulation and hardware-in-the-loop experiments, co-designs the autonomy stack, and streams the measured results back into the model for trade-off analysis and verification, all within one rigorous toolchain. Delivering that integrated, deployable capability, rather than one more model-authoring aid, is what \perfect{}, \tradesx{}, and \veritas{} are built to provide.

\subsection{Trade-off Analysis and V\&V for Autonomous CPS}
Performance benchmarking suites such as RobotPerf~\cite{mayoral2024robotperf} provide reproducible measurement infrastructure for robot autonomy stacks but stop short of SysML integration. Formal verification of autonomy via timed automata (UPPAAL~\cite{behrmann2004tutorial}) and contract-based design~\cite{halder2017formal,pecheur2000verification,d2008survey,drechsler2004advanced} provides sound guarantees on abstracted models but suffers from a fidelity gap with the deployed stack. Hybrid model-based plus data-driven trade-off analysis for autonomous ground vehicles, including risk-sensitive sampling-based planning under Conditional Value-at-Risk (\cvar{}) constraints~\cite{damera2024integrated,enweremRobustStochasticShortestPath2024,kavraki1996probabilistic,lavalle2001rapidly,iagnemmaMobileRobotsRough2004}, has shown order-of-magnitude effort savings relative to pure data-driven evaluation but, again, in isolation from an MBSE context. Most directly comparable are recent digital-engineering frameworks that couple MBSE with digital-twin simulation and requirements-traceable V\&V for autonomous vehicles~\cite{samak-mbse-rdt2025,coleman2025model}; these share \iddmbse{}'s SysML hub and traceability but verify by simulation alone and do not model the learned autonomy stack as a first-class element. MBSE-driven design-space exploration has been formalized with set-based design~\cite{specking2018dse} and model-driven multi-robot exploration on ROS~\cite{Harbin2022Model-driven}, while hardware-centric sensor-suite design, exemplified by Boxi~\cite{frey2025boxi} and task-driven codesign~\cite{Milojevic2025CODEI:}, optimizes payloads empirically from downstream performance rather than from a requirements model. \iddmbse{} differs from all of these by adding proof-level formal verification and the ML/AI stack as a modeled architectural element to the same requirements-driven loop.

\subsection{Bridging the Gap}
The contribution of \iddmbse{} relative to the above is the \emph{integration}: a single methodology that ties SysML structure and behavior models to ML/AI components, a tool chain that connects them to a ROS autonomy stack and a hybrid trade-off engine, and a verification flow that closes the loop between formal model and deployed system, all driven from a single requirements model so that priorities, proofs, and runtime monitors trace to the same source. Each component leverages existing work; the value lies in their composition under one requirements-driven SE discipline (Table~\ref{tab:related}).

%% file: sections/framework.tex
\section{The IDDMBSE Framework}
\label{sec:framework}

\begin{figure}[t]
    \centering
    \includegraphics[width=\linewidth]{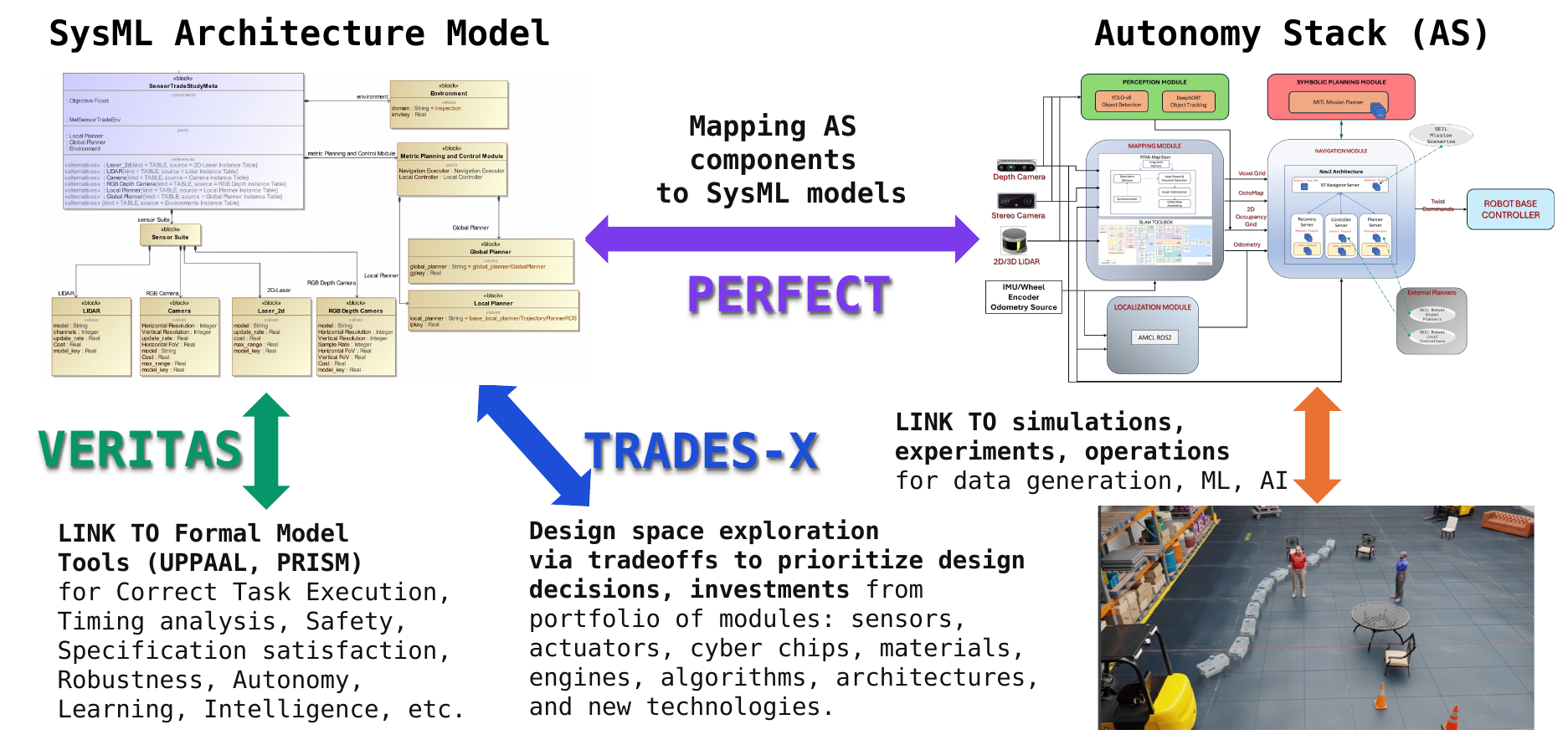}
    \caption{The \iddmbse{} tool chain. \textcolor{perfectPurple}{\perfect{}} maps SysML to ROS\,2 and runs scalable simulations; \textcolor{tradesBlue}{\tradesx{}} performs hybrid model-based plus data-driven design-space exploration; \textcolor{veritasGreen}{\veritas{}} verifies the resulting design via formal, data-driven, and runtime modules. All three share a SysML modeling hub.}
    \label{fig:tools}
\end{figure}

\subsection{Methodology: Data-Driven Augmentation of MBSE}
\label{ssec:methodology}

The organizing principle of \iddmbse{} is requirements-driven and deliberately frugal with data. Because trustworthy simulations and field experiments are scarce for the autonomous robots and missions of interest, the methodology never explores the design space exhaustively; it leans on the model-based prior in two complementary ways. Inexpensive first-principles analysis first prunes the candidates that can be confidently ruled out against requirements, sharply shrinking the set worth deeper study. The scarce empirical budget is then spent only on the survivors, concentrated where the prior is least informative about whether a requirement is met. The requirements repository at the top of the hub (Fig.~\ref{fig:iddmbse_framework_hub}) is thus the entry point, not an afterthought: it both governs what analysis can purge and directs where data must be gathered, so that each V-step invokes data-driven augmentation only where first-principles models leave a requirement uncertain.

The underlying MBSE V-process, which we have refined over a decade of CPS research~\cite{zhou2013cps,spyropoulos2013extending,wang2012integrated,meehan2021slip,sekar2022model} unfolds in six interlocking steps: assess available information, construct \emph{system structure} (what the system consists of), construct \emph{system behavior} (what the system does), map behavior onto structure (which components participate in which behaviors), allocate requirements with traceability, perform trade-off analysis over the resulting design space, and conclude with verification and validation. SysML provides the unifying notation across the first four steps; tools such as Modelica, MATLAB/Simulink, UPPAAL, PRISM, JuMP, Gurobi, and CPLEX provide the analytical machinery for the last two.
\begin{figure}[t]
    \centering
    \includegraphics[width=\linewidth]{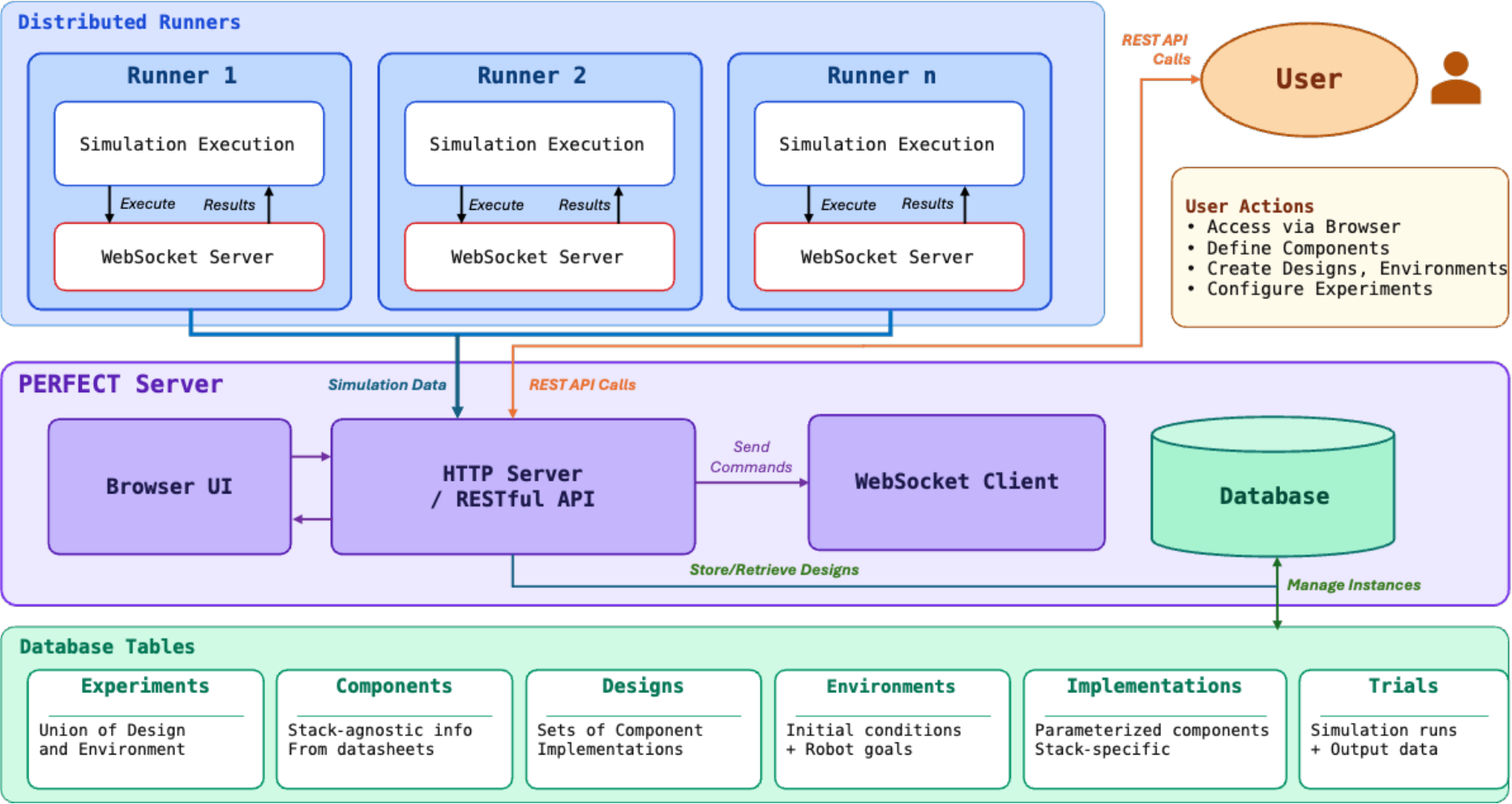}
    \caption{\textcolor{perfectPurple}{\perfect{}} architecture. A central \emph{Server} hosts the component library, an experiment database, and a RESTful plus WebSocket API. Distributed \emph{Runners}, one per compute node, execute simulation jobs and stream results back. The SysML profile maps system blocks to ROS\,2 parametrizations and back.}
    \label{fig:perfect}
\end{figure}

\begin{figure*}
    \centering
    \includegraphics[width=\linewidth]{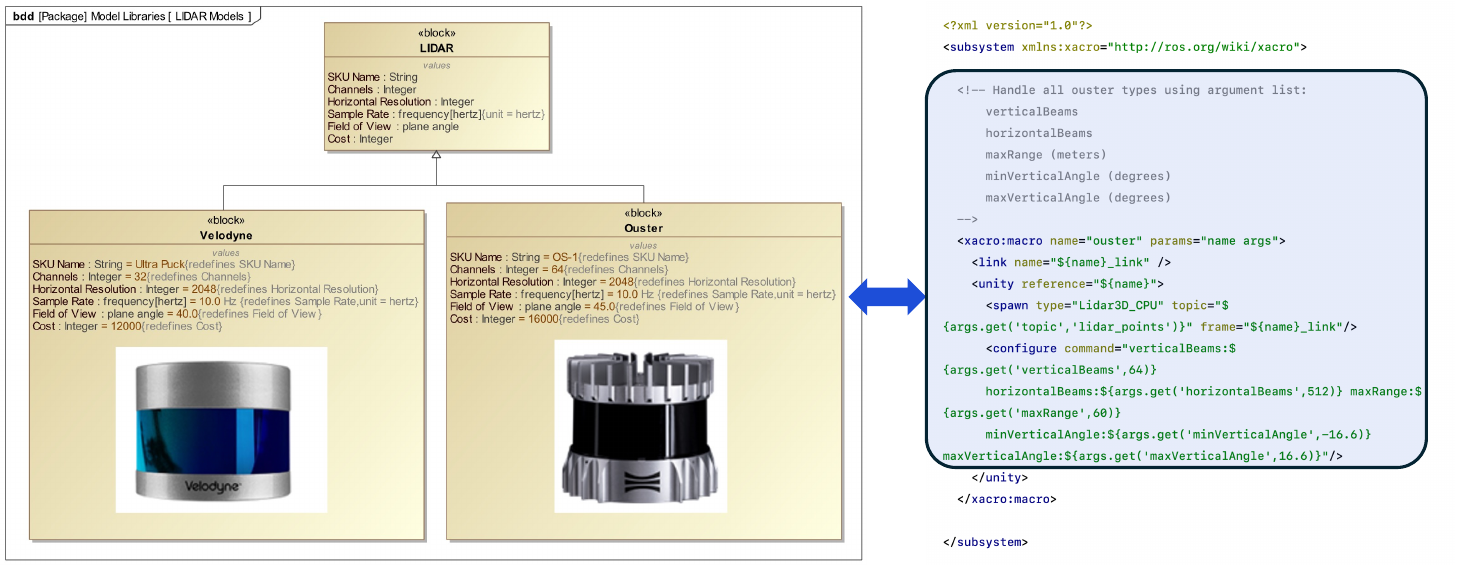}
    \caption{How \perfect{} maps the autonomy stack to SysML. A reusable block library (left) models each sensor as an implementation-agnostic \emph{Component}: a generic LiDAR block specialized into Velodyne and Ouster variants with redefined parametric values (channels, resolution, field of view, cost). The \perfect{} profile binds the chosen block to its stack-specific \emph{Component Implementation} (right), a spawn-and-configure description whose defaults flow from the SysML parametrics. The implementation shown targets a non-ROS engine, illustrating that the binding is agnostic to the runtime backend.}
    \label{fig:mapping}
\end{figure*}

\iddmbse{} adds two structural extensions. The first is a \emph{data-driven loop} at every step: structure and behavior models are augmented by ML/AI components with their own training data and uncertainty profiles; mapping is augmented by data-driven completion of subsystem candidates (an NLP-style analogy in which composable systems are sentences and components are words~\cite{baras2014fresh}); requirements are augmented by data-driven sensitivity ranking that turns their prioritization into a computed quantity~\cite{damera2024integrated}; trade-off analysis is augmented by data-driven evaluation of regions of the design space where first-principles models are weak; and V\&V is augmented by data-driven and runtime verification when formal models are intractable. The second extension is the \emph{autonomy stack as a first-class architectural element}: rather than treat the perception-planning-control stack as an opaque payload sitting on the platform, we model it as a SysML block with structural, behavioral, and parametric ports that bind directly to a ROS\,2 implementation. The three proposed tools described in Fig.~\ref{fig:tools} realize this loop against the shared SysML model: \textcolor{perfectPurple}{\perfect{}} supplies the performance data that closes it, \textcolor{tradesBlue}{\tradesx{}} carries the trade-off augmentation, and \textcolor{veritasGreen}{\veritas{}} carries the verification-and-validation augmentation.


\subsection{PERFECT: SysML-to-ROS2 Performance Evaluation}
\label{ssec:perfect}

\textcolor{perfectPurple}{\perfect{}} (PERFormance Evaluation Composable Toolsuite) addresses the recurring challenge of converting a SysML architecture into a reproducible, scalable performance-evaluation campaign on a ROS\,2 autonomy stack. The tool exposes a SysML profile that binds blocks, parts, and parametric constraints to ROS\,2 nodes, topics, and parameters; once a SysML model is annotated with the profile, \perfect{} synthesizes the ROS\,2 launch configuration, dispatches simulation runs across a distributed compute pool, and aggregates results back into the SysML parametric diagrams for downstream trade-off analysis.

Architecturally (Fig.~\ref{fig:perfect}), \perfect{} consists of a single Server and one or more Runners. The Server hosts a component library (sensor models, planners, controllers, environments), an experiment-results database, and a RESTful plus WebSocket API. Each Runner obtains an experiment definition (task, system, environment parametrization), runs it in Gazebo or Isaac Sim, and streams measurements back to the Server. The novel contribution relative to existing benchmark suites such as RobotPerf~\cite{mayoral2024robotperf} is the SysML linkage: experiment configurations are not free-form scripts but are derived from, and feed back into, the SysML model, preserving the digital thread between architecture and measurement. Concretely, the library is a versioned relational schema that separates an implementation-agnostic \emph{Component}, such as a sensor datasheet, from its stack-specific \emph{Component Implementation} (Fig.~\ref{fig:mapping}), with \emph{Designs}, \emph{Experiments}, and \emph{Trials} composed from these primitives. Retargeting a study to a different autonomy stack or ROS version then touches only the implementation layer and leaves the SysML model intact, and the binding runs over a JSON-over-WebSocket bridge that streams configurations and measurements between model and live stack in real time.


This binding is what makes the rest of \iddmbse{} possible: \tradesx{} and \veritas{} never touch ROS directly but build on the data that \perfect{} returns, so the trade-off loop and the verification layer are each only as scalable as the campaigns \perfect{} can orchestrate. We implement it using commercial SysML\,v1 tooling: \perfect{} drives Magic Systems of Systems Architect through the Magic Model Analyst engine and a SysML-to-MATLAB bridge to reach the live \perfect{} Server, which the v1 tools expose only through brittle, GUI-bound hooks within the SysML architecture trade profiles rather than a clean programmatic interface. We return to the cost of this v1 plumbing, and how an API-first SysML\,v2 dissolves it, in Section~\ref{sec:conclusion}.

\subsection{TRADES-X: Hybrid Design-Space Exploration}
\label{ssec:tradesx}

\textcolor{tradesBlue}{\tradesx{}}  (TRadeoff Analysis and DEsign Space EXploration) addresses the explosion of the design space that occurs when ML/AI components enter the system. Pure model-based optimization fails because not all components admit tractable cost models; pure data-driven evaluation fails because the number of simulation samples required grows combinatorially with design variables.

\begin{figure}[t]
    \centering
    \includegraphics[width=\linewidth]{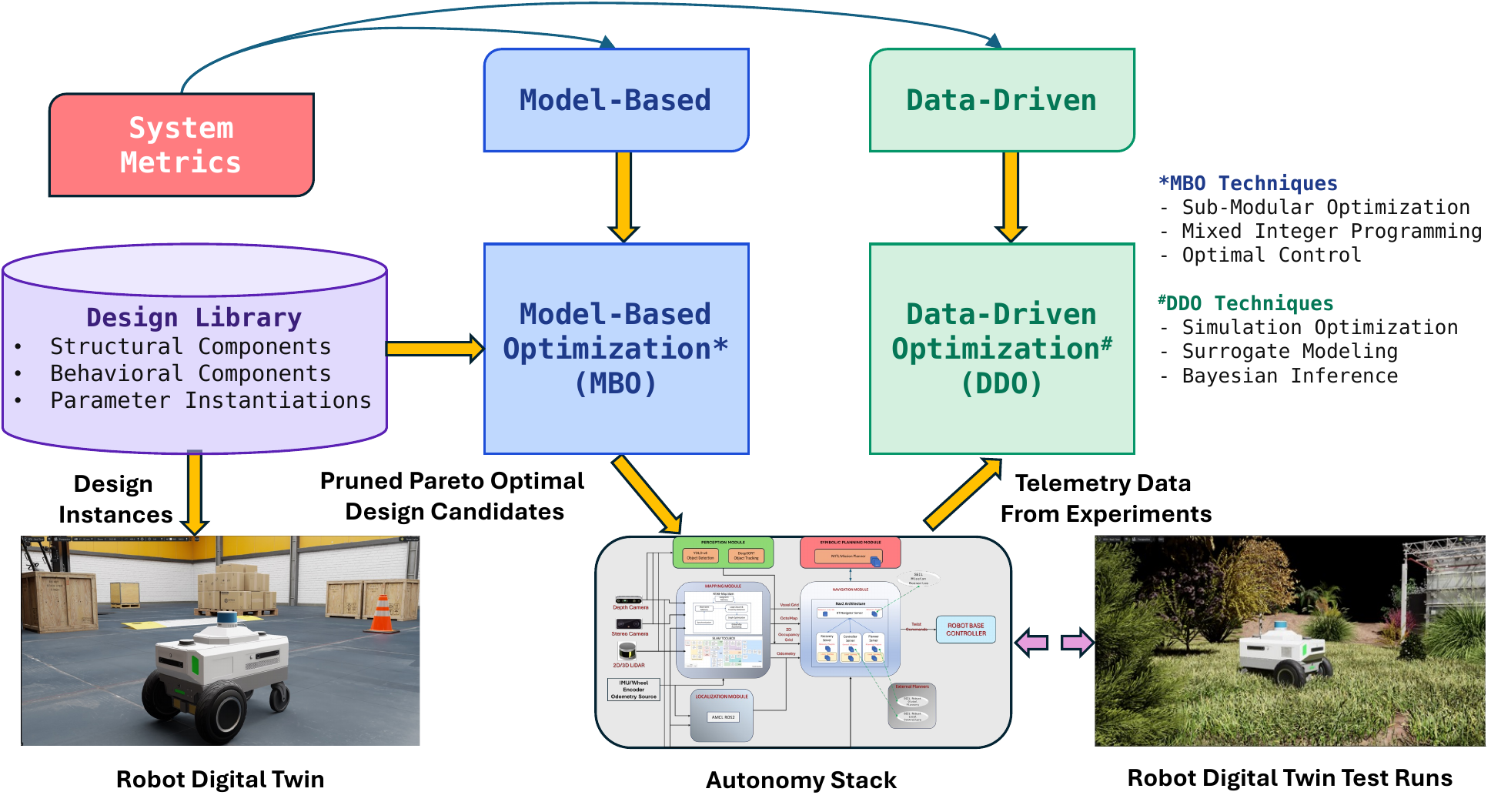}
    \caption{\textcolor{tradesBlue}{\tradesx{}}  decomposes design-space exploration into a Model-Based Optimization (MBO) stage that produces a Pareto-optimal frontier from structured cost models, and a Data-Driven Optimization (DDO) stage that runs simulation experiments on the surviving candidates.}
    \label{fig:tradesx}
\end{figure}

\tradesx{} resolves this by a two-stage decomposition (Fig.~\ref{fig:tradesx}). A \emph{Model-Based Optimization} (MBO) stage formulates a multi-objective offline optimization over the parts of the design space that admit good first-principles models, producing a Pareto-optimal frontier of design candidates. A \emph{Data-Driven Evaluation} (DDE) stage then runs a smaller campaign of statistical experiments on the surviving candidates, using \perfect{} as the execution backbone. A final Multi-Attribute Value Function (MAVF) analysis aggregates the local and global metrics into a single ranking for the stakeholder~\cite{damera2024integrated}. The two-stage structure is analogous to a hierarchical Bayesian estimator: MBO produces a structured prior over feasible designs and DDE updates that prior with empirical likelihoods from simulation. Requirements are partitioned a priori into model-based and data-driven classes, by whether satisfaction can be checked offline or only through experiment, which fixes what MBO proves and what DDE must measure. And \tradesx{} differentiates the metrics with respect to the design parameters, using automatic differentiation to rank requirements by sensitivity and ground the otherwise qualitative assignment of requirement priorities in a quantitative analysis~\cite{damera2024integrated}.

\subsection{VERITAS: Three-Module Verification of Trusted Autonomy}
\label{ssec:veritas}

\textcolor{veritasGreen}{\veritas{}} (VERIfication of Trusted Autonomous Systems) provides the assurance layer of \iddmbse{}. Modern autonomous CPS are complex symphonies of components, resisting single-modality verification: formal model checking requires abstractions that quickly lose fidelity, data-driven testing cannot guarantee absence of faults, and runtime monitoring alone cannot constrain the design. Instead, \veritas{} combines all three, utilizing a variety of model-checking tools and theorem provers to model and verify components, integration with \perfect{} for experimental verification, and the online quantitative monitoring tool RTAMT~\cite{yamaguchi2024rtamt} for runtime verification against safety and liveness specifications.

\begin{figure}[t]
    \centering
    \includegraphics[width=\linewidth]{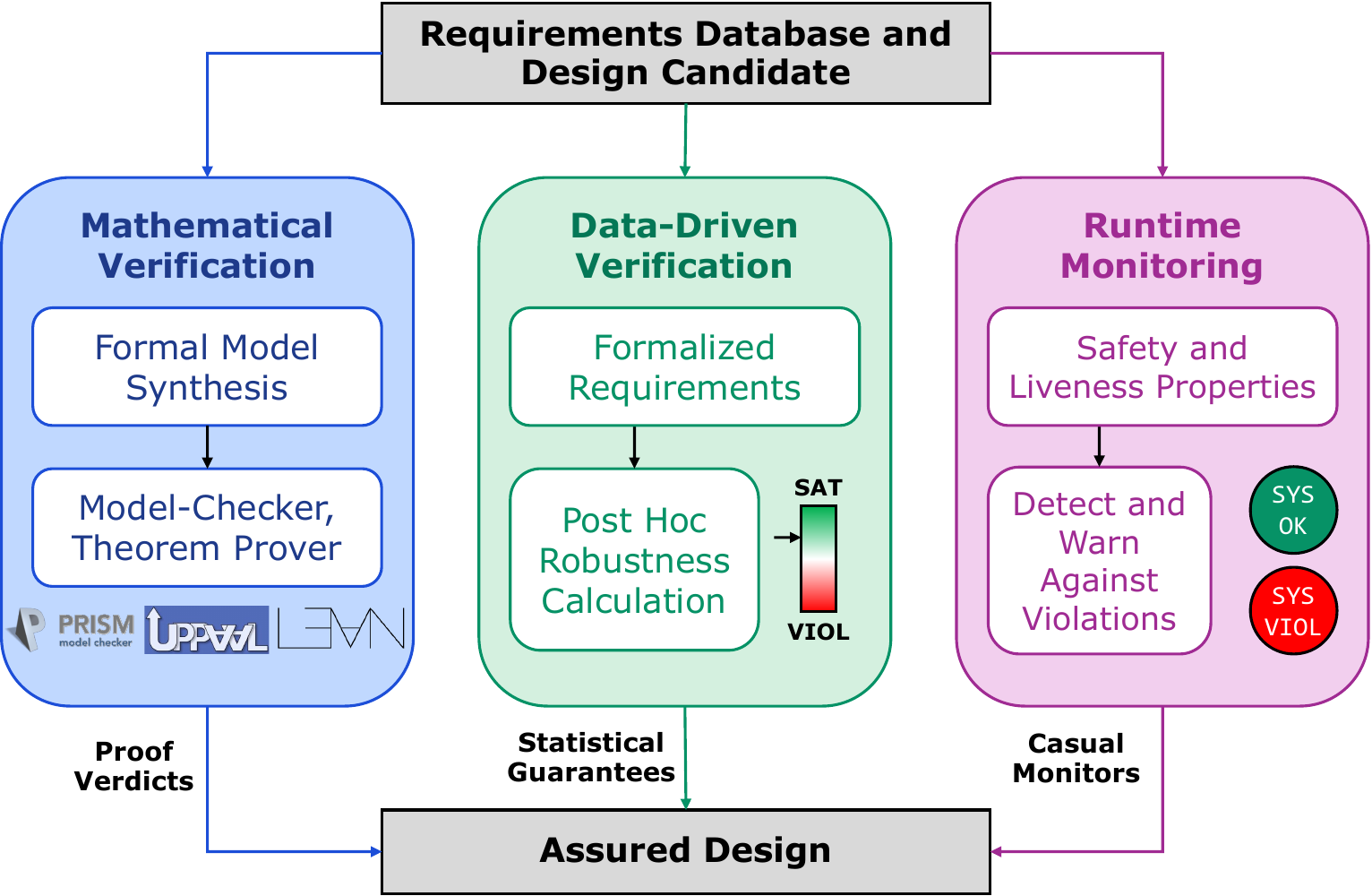}
    \caption{The \textcolor{veritasGreen}{\veritas{}} assurance pipeline. A design candidate and its formalized requirements are certified through three complementary modules: \emph{mathematical verification} (UPPAAL, PRISM, Lean) returns exact proof verdicts; \emph{data-driven verification} computes post-hoc robustness over \perfect{} campaigns for statistical guarantees; and \emph{runtime monitoring} detects and warns against violations online. A candidate becomes an assured design only if it survives all three.}
    \label{fig:veritas}
\end{figure}

The three modules (Fig.~\ref{fig:veritas}) are: (i) a \emph{model-based} module that compiles SysML state-machine and activity diagrams into UPPAAL Networks of Timed Automata~\cite{behrmann2004tutorial,halder2017formal} and discharges proof obligations against contracts derived from requirements diagrams; (ii) a \emph{data-driven} module that uses \perfect{} simulation campaigns to estimate failure rates and confidence bounds for parts of the system where formal models are intractable; and (iii) a \emph{runtime} module that synthesizes observers, typically expressed as Signal Temporal Logic (STL) or Metric Interval Temporal Logic (MITL) monitors, that watch the deployed stack and raise alarms on specification violations. The three modules share a common SysML verification model, so an assurance argument can cite proofs, statistical evidence, and runtime obligations from a single source.

%% file: sections/case_studies.tex
\section{Case Studies}
\label{sec:case_studies}

We demonstrate \iddmbse{} on a Trusted Autonomous Ground Robot (AGR) through two groups of case studies. Section~\ref{ssec:published} spotlights published results that the tool chain enabled, kept deliberately brief since their methods and data appear in the cited works. Section~\ref{ssec:released} develops three demonstrations made available through the open-source \iddmbse{} release, each with a figure, a formulation, and its place in the framework. All exercise the three tools of Section~\ref{sec:framework}, and the contested-terrain Isaac Sim test range introduced in Section~\ref{sssec:range} is the shared environment in which the latter run.

\subsection{Published Demonstrations}
\label{ssec:published}

\subsubsection{Sensor-Suite Selection with \tradesx{}}

Sensor-suite selection for the AGR~\cite{damera2024integrated} exercises \tradesx{} end to end. Over $2^{13}=8192$ candidate suites of 3D LiDAR, 2D laser, RGB, and RGB-D sensors, a model-based optimization stage prunes to $92$ Pareto-optimal candidates by greedy submodular enumeration, \perfect{} simulates only the $7$ survivors, and a Multi-Attribute Value Function analysis returns the recommended design. The campaign collapses from thousands of configurations to single digits: the prune-then-evaluate economy of Section~\ref{ssec:methodology}, realized at the trade-off step.

\subsubsection{Risk-Sensitive Path Planning with \perfect{}}

\begin{figure}[t]
    \centering
    \includegraphics[width=\linewidth]{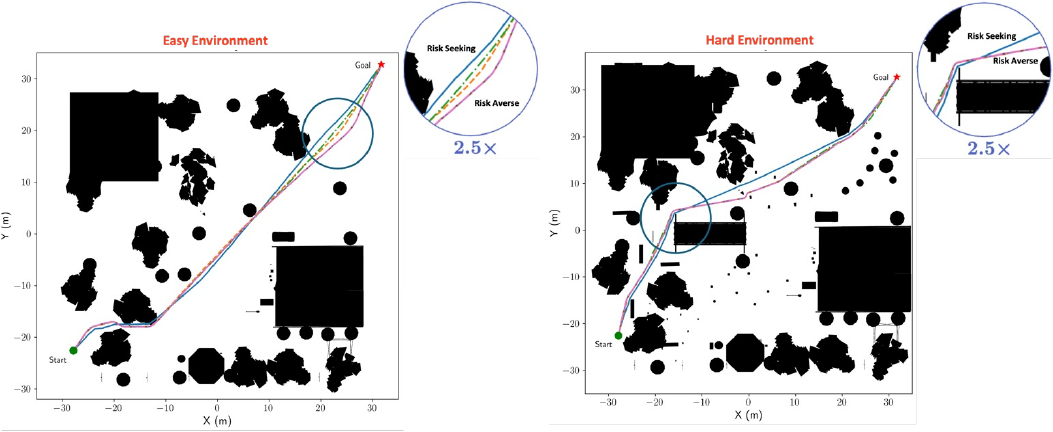}
    \caption{Risk-sensitive planning results for increasing environmental complexity. The risk-averse RA-RRT$^*$~\cite{enweremRobustStochasticShortestPath2024} trades a small path-length premium for routes that stay clear of clutter, an advantage that widens from the easy to the hard environment.}
    \label{fig:rs_planning}
\end{figure}

The risk-sensitive planner RA-RRT$^*$~\cite{enweremRobustStochasticShortestPath2024} casts navigation as a stochastic shortest-path problem and minimizes the sum of per-segment Conditional Value-at-Risk (\cvar{}), retaining the asymptotic optimality of RRT$^*$ while hedging against tail events. Across four noise levels and three risk levels over $50$ runs, it attains lower worst-case path length and markedly fewer planner failures than RRT$^*$ at a modest computation premium, with the risk-averse policy hugging safer routes as the environment hardens (Fig.~\ref{fig:rs_planning}). \perfect{} supplies the distributed campaign behind these comparisons, treating the planner as a behavioral design variable to evaluate at scale.

\subsubsection{Behavior Tree Task Specifications with \veritas{}}

Behavior Trees (BTs) have emerged as a prominent high-level framework for specifying autonomous robot task plans. Concurrently, their widespread adoption has driven a growing emphasis on formal verification to ensure safe and correct operation. To facilitate verification, the \veritas{} tool supports automated model generation~\cite{matheu2025bt2automata} and the synthesis of online monitors~\cite{matheu2025omtbt} directly from BT specifications. Formal models for BTs are then used for verification of task plans and formal control synthesis in UPPAAL. Greedy online monitors for BTs are used for quantitative task monitoring and to provide feedback for learning-based controllers.

\subsection{Demonstrations Released with \iddmbse{}}
\label{ssec:released}

\subsubsection{A Contested-Terrain Test Range in Isaac Sim}
\label{sssec:range}
\begin{figure}[t]
    \centering
    \includegraphics[width=\linewidth]{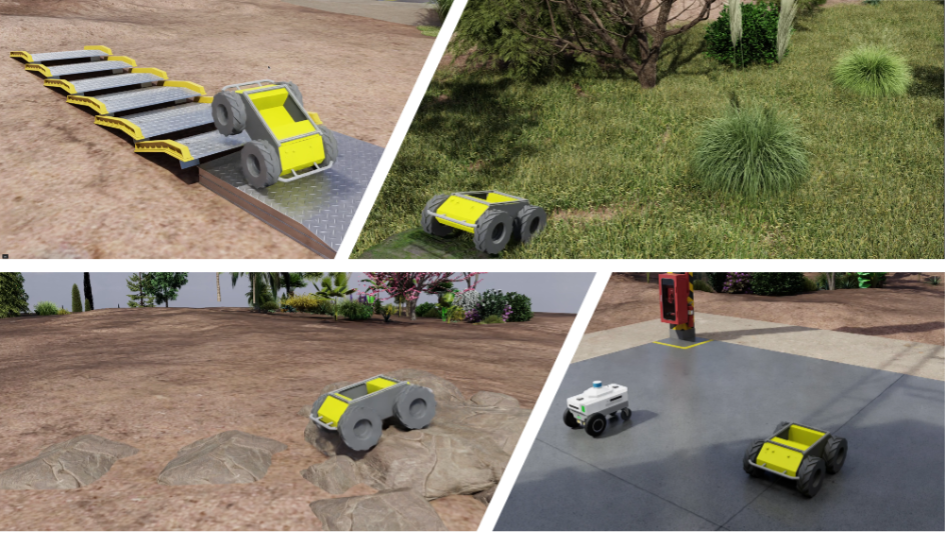}
    \caption{The contested-terrain test range in Isaac Sim, released as USD assets with \iddmbse{}. Terrain slope, surface properties, and obstacle density are exposed as physics parameters that \perfect{} sweeps programmatically for a Design of Experiments.}
    \label{fig:range}
\end{figure}

The demonstrations that follow share a high-fidelity test range we built in NVIDIA Isaac Sim and release as USD assets with the tool chain (Fig.~\ref{fig:range}). It supports multiple AGRs, a full multi-modal sensor payload, and contested off-road terrain with slopes up to $15^\circ$ and obstacle densities swept from $10$ to $80\%$ coverage, all under tunable PhysX physics (surface friction, restitution, and wheel-terrain contact). \perfect{} drives the range through its Python and USD interface: a single Isaac Sim script parameterizes the scene and physics, so a Design of Experiments over terrain conditions becomes a distributed \perfect{} campaign rather than a collection of hand-built worlds. Making the environment a controlled variable is what renders the data-driven loop of Section~\ref{ssec:methodology} reproducible at scale, and the released USD files let others reproduce and extend the campaigns directly.

\begin{figure}[t]
    \centering
    \includegraphics[width=\linewidth]{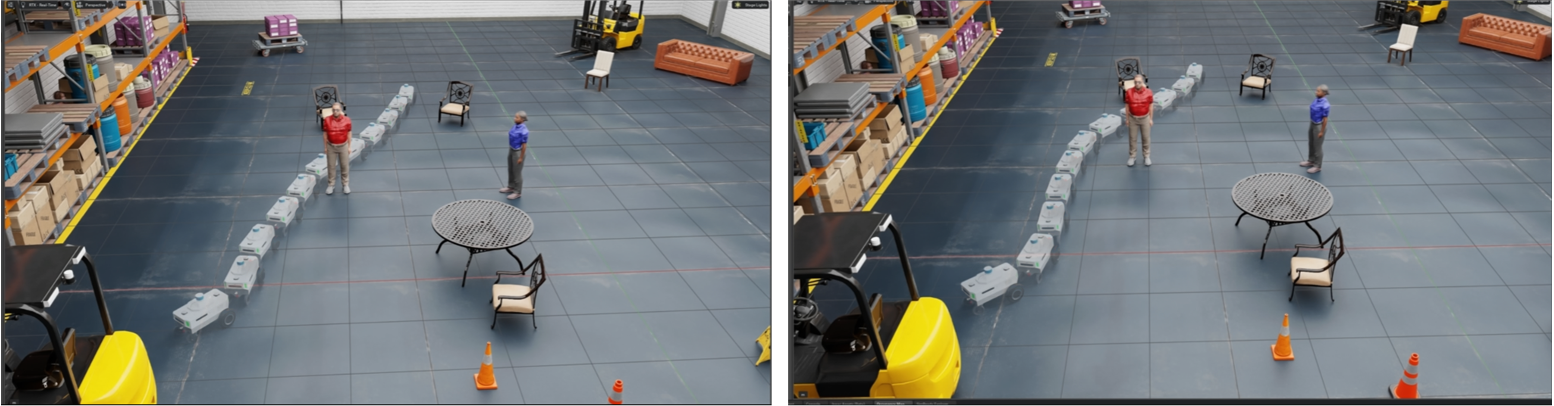}
    \caption{Robust perception via conformal prediction. Realized AGR trajectories without (left) and with (right) conformalized detections: the calibrated obstacle margins steer the planner clear of objects the nominal detector underestimates.}
    \label{fig:conformal}
\end{figure}

\begin{figure*}[htbp]
    \centering
    \includegraphics[width=\linewidth]{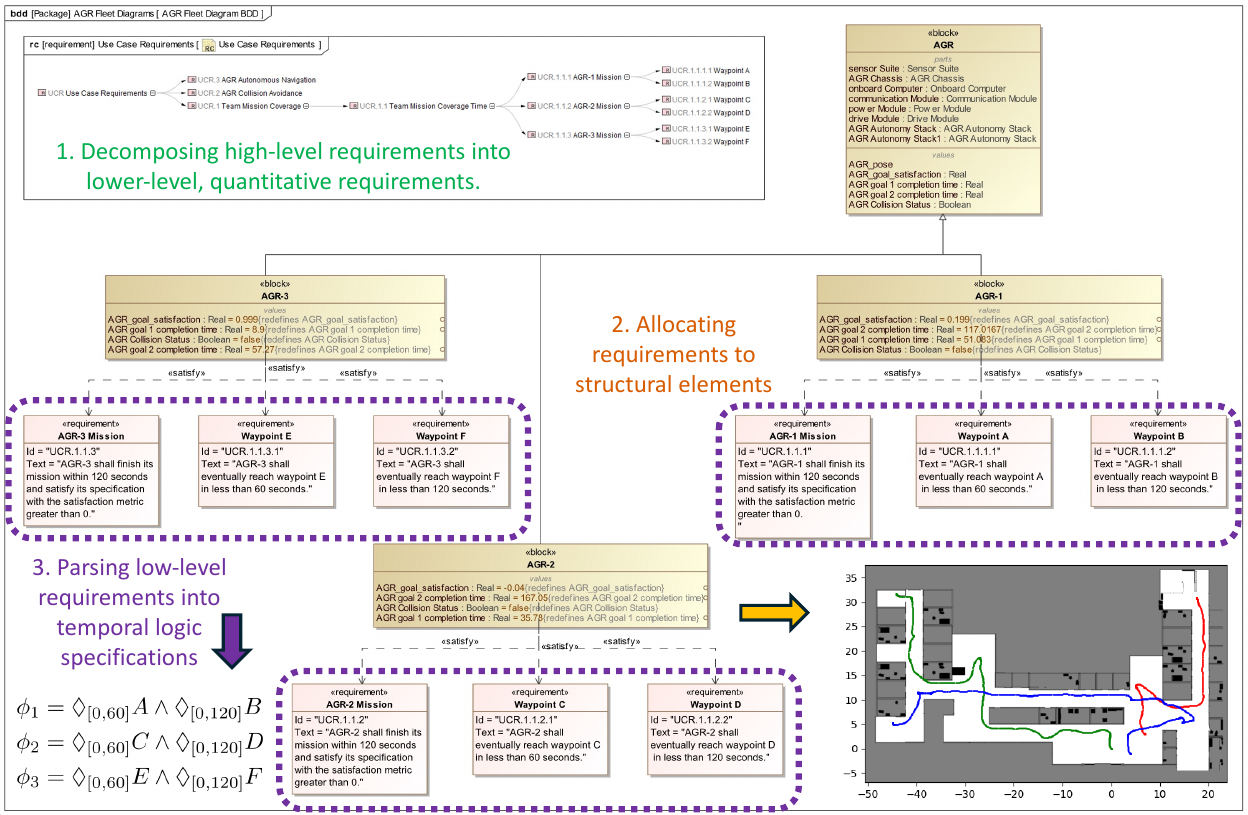}
    \caption{The requirements-driven \iddmbse{} chain for multi-robot coordination. High-level fleet requirements are (1)~decomposed into quantitative per-robot missions, (2)~allocated by \textsf{satisfy} relations onto the structural AGR blocks, and (3)~parsed into reach-avoid STL specifications $\phi_1,\phi_2,\phi_3$. After MILP synthesis and execution (trajectories, lower right), \veritas{} scores each trajectory's STL robustness and writes it back into the model as the block's goal-satisfaction value; robot~2's negative value ($\rho_{\phi_2}=-0.04$) flags its violation while robots~1 and~3 are certified.}
    \label{fig:veritas_demo}
\end{figure*}

\subsubsection{Robust Perception via Conformal Prediction}
\label{sssec:cp}
Learned perception fails silently: a missed or mislocalized detection propagates into an unsafe plan. We harden the perception module with conformal prediction (CP)~\cite{angelopoulos2023conformal}, which wraps any detector in a distribution-free guarantee. Given a calibration set, CP constructs a prediction region $R_\alpha(x)$ around a detector output $f(x;\theta)$ such that
\begin{equation}
    \mathbb{P}\!\left(f(x;\theta)\in R_\alpha(x)\right) \ge 1-\alpha,
    \label{eq:cp}
\end{equation}
for a user-chosen miscoverage level $\alpha$. We apply~\eqref{eq:cp} to object detection, inflating each bounding box to its $(1-\alpha)$ conformal region before it reaches the planner, so that collision avoidance carries an assurance which survives the closed-loop distribution shift the planner itself induces, following the end-to-end navigation-safety construction of~\cite{mei2026perceive} (Fig.~\ref{fig:conformal}). The point we stress is not the calibration theory, which is developed in~\cite{mei2026perceive}, but that the \iddmbse{} tool chain is what made it executable on our AGR. Conformal calibration is only as sound as its calibration data, which must be ground-truth-labeled and drawn from the closed-loop states the robot will actually visit; \perfect{} produces exactly that, running the full perception-and-planning stack across the contested-terrain range (Section~\ref{sssec:range}) to generate labeled, closed-loop trajectories at scale and re-generate them under deliberate distribution shifts in lighting, clutter, and terrain. \tradesx{} then treats the coverage level $1-\alpha$ as a design variable, trading it against navigation conservativeness to fix the operating point where larger obstacle margins buy safety without stalling the robot.

\subsubsection{Assured Multi-Robot Coordination}
\label{sssec:assured-ma}
The framework scales from one robot to a team, and this case ties the whole requirements-driven chain together (Fig.~\ref{fig:veritas_demo}): high-level fleet requirements are decomposed into quantitative per-robot missions, allocated by \textsf{satisfy} relations onto the structural AGR blocks, and parsed into the reach-avoid Signal Temporal Logic (STL) specifications $\phi_1,\phi_2,\phi_3$ shown there. We pose the joint navigation task as a Mixed-Integer Linear Program: a big-M encoding turns the conjunction of linearized dynamics and STL-satisfaction constraints over the halfspace-representation environment into a MILP solved to global optimality with off-the-shelf solvers. \veritas{} then closes the loop, scoring each executed trajectory against its specification by the quantitative STL robustness $\rho$ (positive if and only if satisfied) and writing the verdict back into the AGR block as its goal-satisfaction value: $\rho_{\phi_1}=0.199$ and $\rho_{\phi_3}=0.999$ certify robots~1 and~3, while $\rho_{\phi_2}=-0.04$ isolates robot~2's slight violation, the data-driven verification module catching what synthesis alone would miss. \tradesx{} explores the communication-topology and task-allocation trade-offs over the team's collaboration-communication-information multigraph~\cite{baras2014fresh}.

%% file: sections/conclusion.tex
\section{Conclusion and Future Work}
\label{sec:conclusion}

We presented \iddmbse{}, a requirements-driven, data-driven augmentation of the MBSE V-process for Trusted Autonomous CPS, realized as three interoperable tools on a shared SysML hub: \perfect{} for SysML-to-ROS performance evaluation, \tradesx{} for hybrid design-space exploration, and \veritas{} for formal, data-driven, and runtime verification. Across the robot development lifecycle, sensor-suite selection, path-planning evaluation, conformalized perception, and multi-robot coordination, the same requirements thread carries from model to deployed, verified system, with assurance that cites proof, statistical evidence, and runtime robustness from one source.

The chief limitation is the SysML\,v1 substrate. As Section~\ref{ssec:perfect} noted, binding the model to a live stack today means coercing it through Magic Model Analyst and a SysML-to-MATLAB bridge that is brittle, GUI-bound, and vendor-locked. Our ongoing work re-casts \iddmbse{} on SysML\,v2~\cite{omgSysMLv2}, whose textual, API-first surface dissolves that plumbing: a standardized JSON API exposes the model as a queryable graph that \perfect{} drives directly, KerML's first-class part and variant semantics replace custom profiles, and the textual form places the digital thread under version control.

That API also opens an \emph{agentic} \iddmbse{}, in which each V-step is delegated to an AI agent acting on the model graph through governed runtimes such as OpenClaw/NemoClaw, so that ML/AI enters the loop with traceability rather than smuggling unverified content into the digital thread. We will release the \iddmbse{} tool chain and its contested-terrain test range to support this transition.